%% file: main.tex
\documentclass[10pt,twocolumn,letterpaper]{article}

\usepackage[pagenumbers]{iccv} 

\usepackage{times}
\usepackage{epsfig}
\usepackage{graphicx}
\usepackage{amsmath}
\usepackage{amssymb}
\usepackage{gensymb}

\usepackage{float}
\usepackage{multirow}
\usepackage{array}
\usepackage{tabularx}
\usepackage{adjustbox}
\usepackage{booktabs}
\usepackage{algorithm}
\usepackage{algpseudocode}

\newcolumntype{C}[1]{>{\centering\arraybackslash}p{#1}}

\definecolor{iccvblue}{rgb}{0.21,0.49,0.74}
\usepackage[pagebackref,breaklinks,colorlinks,allcolors=iccvblue]{hyperref}

\def\confName{ICCV}
\def\confYear{2025}

\title{Inside Knowledge: Graph-based Path Generation with Explainable Data Augmentation and Curriculum Learning for Visual Indoor Navigation}

\author{
Daniel Airinei$^{1}$ \quad Elena Burceanu$^{2}$ \quad Marius Leordeanu$^{1,2, 3}$\\
$^{1}$ National University of Science and Technology POLITEHNICA Bucharest, Romania \\ $^{2}$ Bitdefender \quad  $^{3}$ Institute of Mathematics ”Simion Stoilow” of the Romanian Academy, Romania \\
{\tt\small   constantin.airinei.stud.acs.upb.ro \quad eburceanu@bitdefender.com \quad leordeanu@gmail.com}
}
\begin{document}
\maketitle
\input{sec/0_abstract}    
\input{sec/1_intro}

\input{sec/2_related_work}
\input{sec/3_proposed_method}

\input{sec/4_experiments}
\input{sec/5_conclusions}

\section*{Acknowledgements}
This work is supported in part by projects “Romanian Hub for Artificial Intelligence - HRIA”, Smart Growth, Digitization and Financial Instruments Program, 2021-2027 (MySMIS no. 334906) and "European Lighthouse of AI for Sustainability - ELIAS", Horizon Europe program (Grant No. 101120237).

{\small
\bibliographystyle{ieee}
\bibliography{egbib}
}

\end{document}

%% file: sec/0_abstract.tex
\begin{abstract}
Indoor navigation is a difficult task, as it generally comes with poor GPS access, forcing solutions to rely on other sources of information. While significant progress continues to be made in this area, deployment to production applications is still lacking, given the complexity and additional requirements of current solutions.
Here, we introduce an efficient, real-time and easily deployable deep learning approach, based on visual input only, that can predict the direction towards a target from images captured by a mobile device. Our technical approach, based on a novel graph-based path generation method, combined with explainable data augmentation and curriculum learning, includes contributions that make the process of data collection, annotation and training, as automatic as possible, efficient and robust. On the practical side, we introduce a novel large-scale dataset, with video footage inside a relatively large shopping mall, in which each frame is annotated with the correct next direction towards different specific target destinations.  
Different from current methods, ours relies solely on vision,
avoiding the need of special sensors, additional markers placed along the path, knowledge of the scene map or internet access. We also created an easy to use application for Android, which we plan to make publicly available. We make all our data and code available along with visual demos on our project site\footnote{\url{https://sites.google.com/view/indoor-navigation-by-vision/}}.
\end{abstract}

%% file: sec/1_intro.tex
\section{Introduction}
Navigation is an essential part of human life, and an increasingly important domain, given the advancements in robotics, and the need for self-navigating cars and robots. Large indoor structures, such as shopping malls or subway stations can be confusing for people who do not frequent them, and wayfinding inside them can be a tedious task, which can be eased using approaches such as the one presented in this paper. Indoor navigation is also a central area of research in aiding people with visual impairments navigate public places with which they are unfamiliar.

Indoor navigation is markedly distinct from its outdoors counterpart, where both humans and machines would rely on GPS for broad localization, and on vision for the actual maneuvering through the environment. Indoor navigation comes with the challenge of having weak access to GPS service\cite{mikkel}, due to the building structure. The solution must now rely on vision only, for both local movement (avoiding obstacles, knowing when you are in an intersection), as well as global direction (knowing which way to go in an intersection). However, this is rarely the case for indoor solutions, as it is common to make use of additional equipment, such as special mapping of the scene, setting up markers and special indicators across it. However, relying on custom changes to the scene is undesirable, as a guiding system would ideally be independent from pre-deployed infrastructure\cite{hashish}. Collection of LiDAR, WiFi, BlueTooth fingerprints and other similar data sources, are also commonly used approaches which will be discussed in depth in Section \ref{sec:related_work}. Additionally, navigation can be achieved by computing direction from camera feed data alone, by methods that localize the user on a known map, and computing orientation based on that information. We consider our method to stand out among these, as we achieve navigation without global localization on a map, using 2D images alone, collected from a consumer grade mobile device.
We train a ConvNet to predict the walking direction towards a target from a predefined set, in real-time, based on the camera feed of a mobile device.

\noindent We make the following contributions:
\begin{enumerate}
\item \textbf{Automatic graph-based generation of novel training paths.} We introduce an efficient graph-based method to generate new indoor paths between all possible start and end points, which offers a full set of possibilities for training by combining automatically segments of real video footage.
\item \textbf{Improved learning strategies using explainability and curriculum learning.} Using the Grad-CAM algorithm for explaining which parts of input are responsible for errors, we design effective partial masking of the input at training time, to improve robustness. Combined with curriculum learning, in which hard training cases are added during the last stage of training, these strategies help the model overcome most of its cases of failure. 
\item \textbf{Novel indoor navigation dataset with automatic direction annotation.} We recorded data in a real-life scenario, by walking around and recording video footage in a relatively large shopping mall, at different times of the year and with different levels of agglomeration. We compute the true ground truth direction on the world map, with an efficient vision-only method from the observed optical flow and monocular depth, both automatically estimated. 
\end{enumerate}

%% file: sec/2_related_work.tex
\section{Related Work}
\label{sec:related_work}
It is common for research regarding indoor navigation to propose methods that are ultimately aimed at being incorporated into a commercial product that can help people, especially those with impaired vision, to navigate their surroundings. As such, a lot of studies also propose  a mechanism for user guidance, such as audio instructions or a graphic interface. We also do this through visual cues on the screen in the form of a compass. However, in this section, we will focus on the related work mostly from the perspective of the navigation mechanism, since that is the area to which we bring our main contribution.
Outdoor navigation is a related field from which we can import relevant approaches and techniques. Many studies concern outdoor navigation in the context of car driving, either for humans\cite{Leordeanu, Vu, amini} or autonomous driving\cite{barua,kumar2022vision,Do,Lopez}. Other outdoors related studies focus on pedestrians\cite{ruotsalainen, treuillet} and UAV's\cite{arafat,zhang2011}. However, our focus is placed on indoor scenarios, where conditions differ and new techniques must be developed. 
There are a large number of studies regarding indoor navigation, and several surveys\cite{El-Sheimy2021, 8593096, Huang2010, KHAN202224, s20092641} that describe them. These surveys generally differentiate methods by a few fundamental differences, such as type of sensors used, requirement of scene maps, requirement of special markers, etc. We will further discuss these categories and position our approach among them.

\textbf{Navigation based on special sensors}. As mentioned, most indoor navigation relies on special technologies, such as RFID, WiFi networks, BlueTooth, infrared, inertial sensors, accelerometers, and others. While they are effective, they are considerably different from our approach, and we can consider them all to be non-vision based methods.

Of these, we note some that localize the user globally\cite{Loomis2006,Beydoun2008,Ando2009} on a known map. Others\cite{paredes, drones6080208} use sensors without positioning the client on an available plan of the scene, more akin to our method. Zheng \textit{et al.}\cite{Zheng2017} create a mobile application that navigates the user based on pre-recorded paths. They record WiFi and motion data, and match it algorithmically with the historical data, determining which way a user should be going. Their application displays a static image of the correct path to the user, who then has to orient himself to identify it in his surroundings. 

\textbf{Navigation based on vision}.
Here we find it relevant to make a distinction between solutions that use consumer grade devices, like ours, and ones that use special equipment that can capture stereo images, or 3D images. Of the latter, there are methods such as that described by Teng \textit{et al.}\cite{10.1145/3216722}, where the scene is mapped using a 3D camera, and the location of the users is then inferred by using a point cloud similarity matching algorithm. RGBD cameras provide depth data that can be used to perform operations such as SLAM and obstacle detection\cite{LEE20163, kanwal, 8370712, jin, Endres}.  

Of the methods that use mobile camera footage, a common approach is using preset markers such as color codes or QR codes on the scene\cite{Manduchi2010, Zeb2014, Chang2008, khan2019generic, botta2020performance, MEKHALFI2016129, anup}, which increase the navigation reliability at the cost of an increase in deployment difficulty.

Those that do not use markers, rely on scene recognition to place the user on a map\cite{royer2007monocular}. Karkar \textit{et al.}\cite{Karkar2021} propose an application in which the user sends images of the scene to a server, that uses deep learning to infer the coordinates of the user, and sends back the position where the images were taken. Fusco \textit{et al.}\cite{Fusco2020} localize a user based on exit signs captured within images captured by a phone.  

\textbf{Main novelty of our approach:} it is self-contained, not requiring access to external devices or setup at test time, or access to a building map. We implemented it on a consumer-grade smart phone, requiring only access to its camera, thus being an affordable, easily deployable solution. In this category, methods such as the one described by Li \textit{et al.}\cite{li2021multi} can also be included, but are primarily concerned with robot navigation within a room, as opposed to wayfinding in a large building, a task fundamentally different from the one we work with. Padhy \textit{et al.}\cite{padhy2018deep} propose a similar system that allows a drone to autonomously traverse indoor corridors by classifying frames with flight commands, but mostly for low-level maneuvering purposes, not for reaching a target destination.

%% file: sec/3_proposed_method.tex
\section{Proposed method}
For helping a person to navigate indoors, it is sufficient to give at every moment and in real-time, one of several
high-level directions (e.g. forwards, left, right, slight-right, turn around), which will guide that person towards a desired target location.
We approach this task as a classification problem, where each direction belongs to a set of discrete classes. Given a video stream captured in real-time with a mobile device, the system will learn to automatically predict, based on visual input alone, one of the direction classes - without any additional sensor information (e.g. GPS, depth sensor, access to the Internet) or knowledge of the map.

\begin{figure}[!h]
    \centering
    \includegraphics[width=0.9\columnwidth]{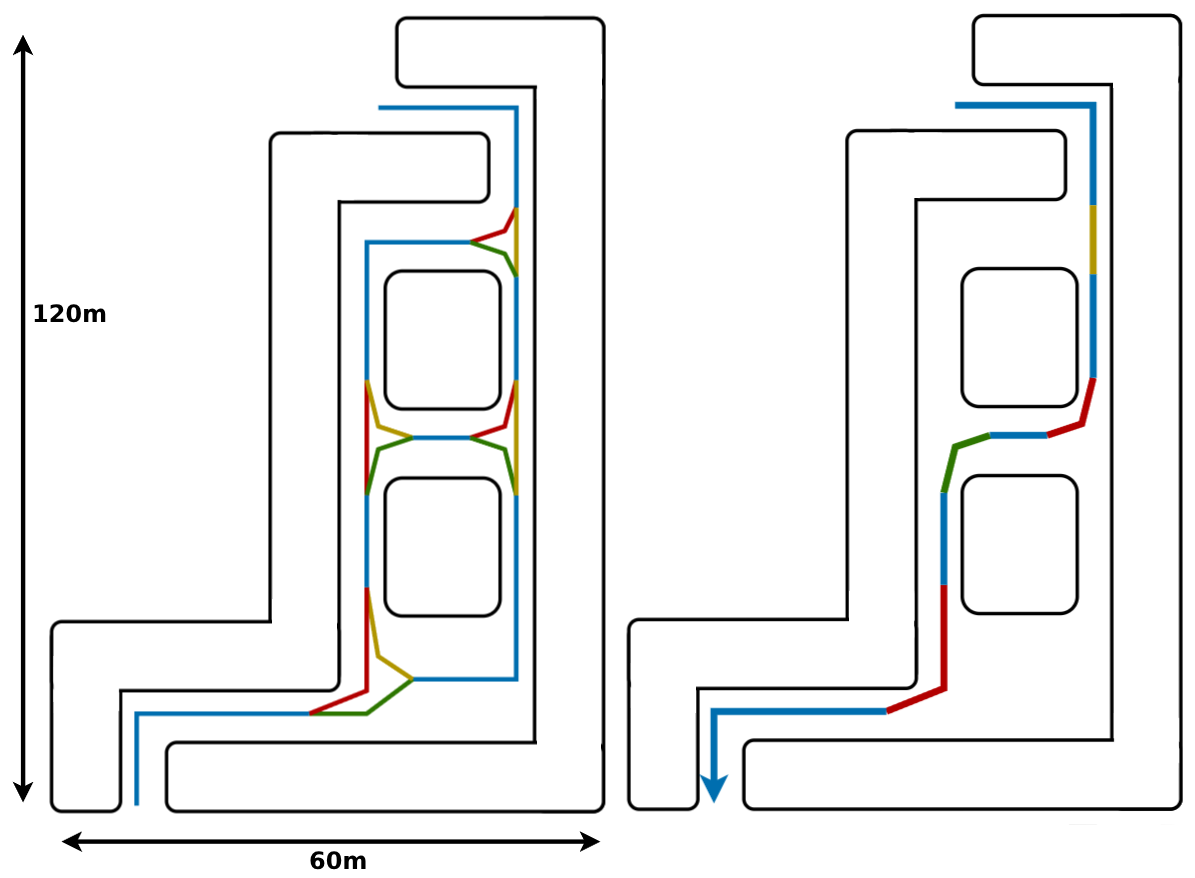}
    \caption{Dataset location floor plan, with visualization of annotated segments (\textbf{left}) and generation of a new path that was not ever fully filmed (\textbf{right}).}
    \label{fig:cora}
\end{figure}

\subsection{Data acquisition and preparation}
There is currently no published dataset, as required by our approach, which is why we had to collect one.
Previous works use simulated environments for similar tasks, but real data is actually needed in order to train, test and develop a robust, real-world application, which is our end practical goal. Thus, we chose an appropriate indoor environment, a medium-sized shopping mall, which is fitted for sufficient data collection for extensive training and testing, during different times of the year, of an indoor navigation system with real practical value.
The video footage is recorded with a mobile phone, held at shoulder level, while walking towards different desired destinations, with levels of precision that would be expected of an end user. In some videos the phone is more tilted than in others and the footage usually suffers from shakiness when taking every step. 
The filming location (general outline can be seen in Figure~\ref{fig:cora}) spans roughly 120 meters in length and 60 meters in width, features 4 intersections and a large number of stores along the corridors that can be used as target destinations. In total, we have around 3 hours of footage filmed along corridors of the mall, with most corridors being visited at least 4 times. To be able to guide a user from one point to another, we have to be able to include all frames of that path at training time, however we found it is not feasible to walk all possible combinations of points in the store.  
Instead, we exhaustively record footage that contains all possible segments of the mall, annotate them based on a topological map (Figure~\ref{fig:cora}), where intersections are nodes, corridors are edges, and we differentiate ways in an intersection based on the starting and destination corridor. We can then generate any custom path in the mall using these segments, or automatically generate training data as described in algorithm~\ref{alg:generate_paths}.

\begin{algorithm}[!h]
\caption{Generate synthetic navigation paths via graph-based video recombination}
\label{alg:generate_paths}
\begin{algorithmic}[1]
\Require Graph $G = (V, E)$ where $V$ are intersections and exits (nodes) and $E$ are corridors (edges), annotated video segments for all consecutive triplets, edge weights $w(e)$ as video length (in frames)
\Ensure A valid synthetic path from a random start to a random goal
\State Sample random start node $v_s \in V$
\State Sample random goal node $v_g \in V$
\State Compute shortest path $P = \texttt{Dijkstra}(G, v_s, v_g)$ using edge weights $w(e)$ (minimize total length)
\State Initialize path triplets list $\mathcal{T} \leftarrow []$
\For{each consecutive triple $(e_i, v_i, e_{i+1})$ along path $P$}
    \State Retrieve annotated triplet video segment for $(e_i, v_i, e_{i+1})$
    \State Append triplet $(e_i, v_i, e_{i+1})$ to $\mathcal{T}$
\EndFor
\State \Return $\mathcal{T}$ as the synthetic path sequence
\end{algorithmic}
\end{algorithm}
\subsection{Camera motion extraction}
For training, we need the ground-truth 3D motion of the camera with respect to the world, which is not trivial to obtain from vision alone, without additional motion sensors (which are often noisy). To get the direction class, during training, we are only interested in the camera rotation around the Y axis, assuming the phone is held almost vertically.
To obtain a precise estimation of this rotation we propose a fast and efficient vision-only solution, which combines the observed optical flow, monocular depth estimation and actual 3D camera motion, based on results from visual-based robotics \cite{corke2011robotics}, in combination with state of the art optical flow and monocular depth estimation deep networks. Note that this procedure is needed only during the training process and could be performed offline on any system with a GPU. 
The equation that relates pixel coordinates with respect to the principal point, camera motion, optical flow, and depth is:

\[
\setlength{\arraycolsep}{0.5pt} 
\begin{pmatrix}
\dot{u} \\ \dot{v}
\end{pmatrix}
=
\begin{pmatrix}
-\frac{f}{\rho Z} & 0 & \frac{u}{Z} & \frac{\rho u v}{f} & \frac{f^2 + \rho u^2}{f} & \frac{\rho v}{f} \\
0 & -\frac{f}{\rho Z} & \frac{v}{Z} & \frac{f^2 + \rho v^2}{f} & -\frac{\rho u v}{f} & -\frac{\rho u}{f}
\end{pmatrix}
\begin{pmatrix}
v_x \\ v_y \\ v_z \\ \omega_x \\ \omega_y \\ \omega_z
\end{pmatrix}
\tag{1} \label{eq:motion}
\]

\begin{figure}
    \centering
    \includegraphics[width=0.8\columnwidth]{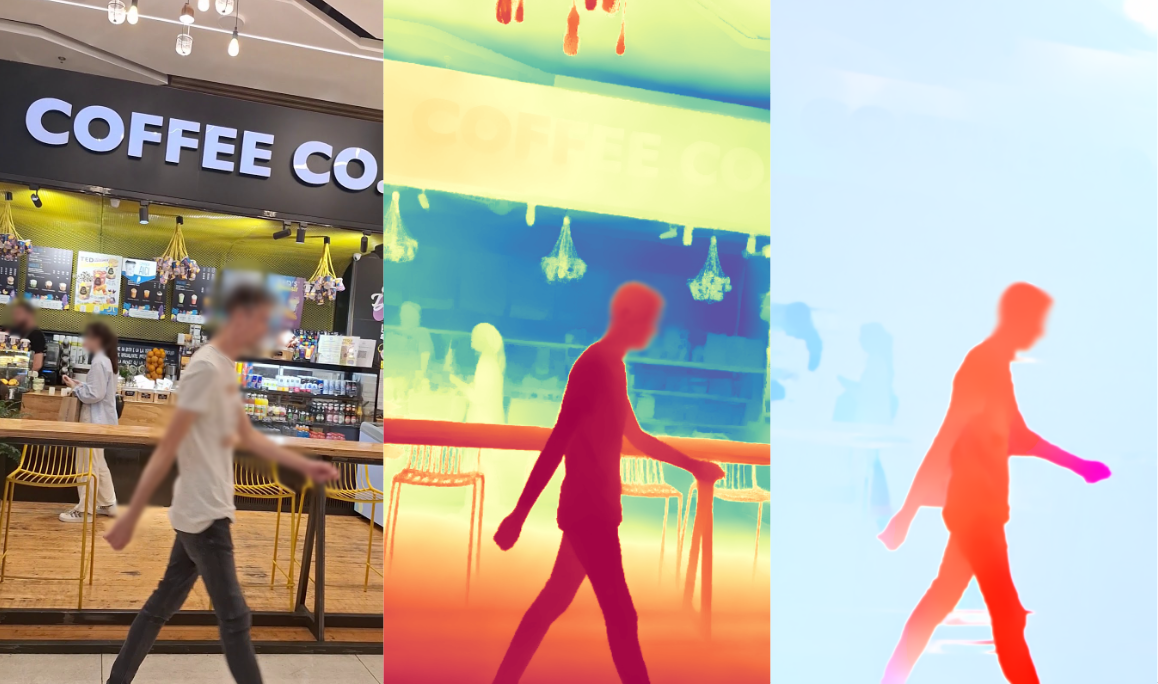}
    \caption{Original frame (\textbf{left}). Depth estimation (\textbf{center}). Optical flow (\textbf{right})}
    \label{fig:depth}
\end{figure}

The left-hand side of the equation describes optical flow, which can be obtained from the video footage. We choose a pretrained deep learning model\cite{dosovitskiy2015flownet} to do so. The Z parameter is the depth of the scene over the pixel coordinates, which can also be obtained using a pretrained model\cite{viola2024marigold}. An example of the output of these models alongside the original frame can be visualized in Figure~\ref{fig:depth}. The rest of the parameters in the second factor of the equation are the actual pixel coordinates and the intrinsic parameters of the recording device. These are available online for most consumer devices. The only unknown in the equation is the third factor which represents all 6 components of the camera movement (translation and rotation on 3 axis each). We are only interested in $ \omega_y $, which will always correlate with the user's orientation throughout the path.

To isolate the rotational component around the Y-axis, from the combined motion represented in Equation~\eqref{eq:motion}, we adopt a least squares estimation approach from many $N$ image samples. It can then be interpreted as a linear system of the form:
\[
\dot{\mathbf{u}}_i = A_i \mathbf{x},
\]
where $\dot{\mathbf{u}}_i = (\dot{u}_i, \dot{v}_i)^\top$ denotes the optical flow vector at the $i$-th pixel, $A_i \in \mathbb{R}^{2 \times 6}$ is the pixel-wise motion matrix determined by depth $Z_i$, image coordinates $(u_i, v_i)$, and intrinsic parameters, and $\mathbf{x} = (v_x, v_y, v_z, \omega_x, \omega_y, \omega_z)^\top$ is the unknown camera motion vector.

By aggregating $N$ such measurements across randomly sampled pixels, we construct the stacked system:
\[
\mathbf{b} = A \mathbf{x},
\]
where $A \in \mathbb{R}^{2N \times 6}$ and $\mathbf{b} \in \mathbb{R}^{2N}$ are the concatenations of all $A_i$ and $\dot{\mathbf{u}}_i$ respectively. Given that there are 6 unknowns and each pixel point results in 2 equations, we need at least 3 points to approximate a solution. We solve for $\mathbf{x}$ using the standard least squares solution, by randomly sampling 200 points in the frame for robustness of approximation:
\[
\hat{\mathbf{x}} = \arg\min_{\mathbf{x}} \|A\mathbf{x} - \mathbf{b}\|^2 = (A^\top A)^{-1} A^\top \mathbf{b}.
\]
From the estimated motion vector $\hat{\mathbf{x}}$, we extract $\omega_y$ as the component of interest, which corresponds to the yaw rotation of the camera and hence the user's heading direction.

\subsection{Motion data interpretation}
\begin{figure}[ht]
    \centering
    \includegraphics[width=0.8\columnwidth]{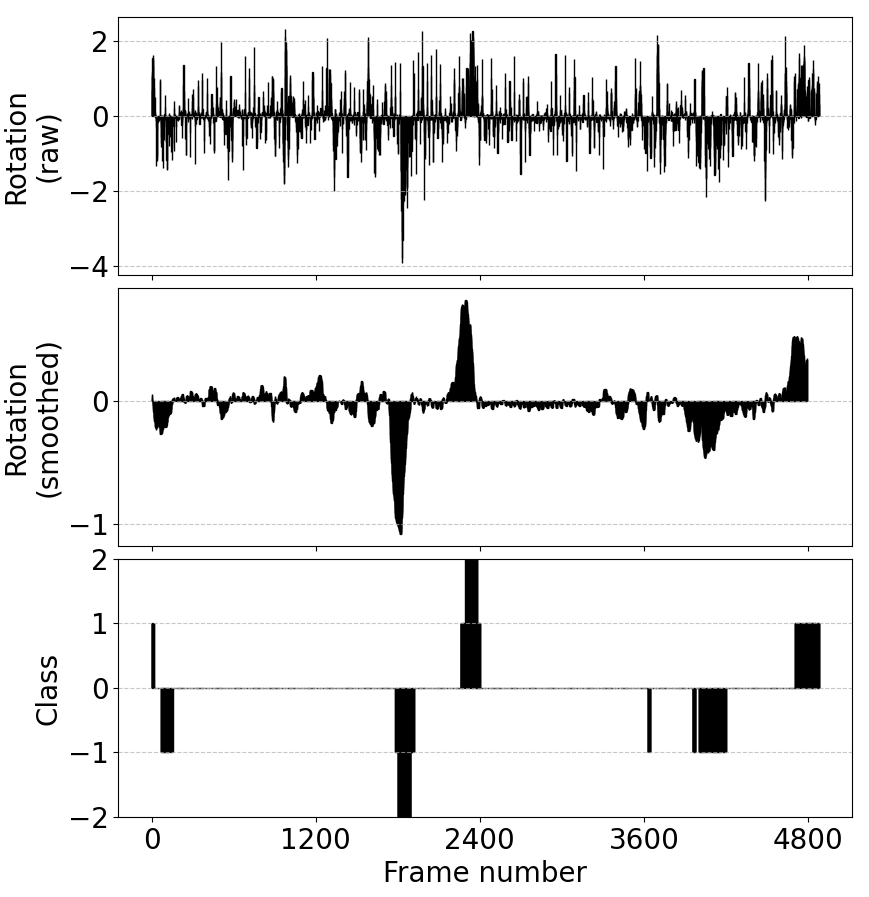}
    \caption{Plots of motion data, where the horizontal axis represents the frame number of the video and vertical axis the rotation component. Raw motion data (\textbf{top}), smoothed values (\textbf{center}) and conversion into 8 possible discrete values (\textbf{bottom}). The 5 peaks visible in the bottom plot correspond to 5 turns in the building in the video. The 3 classes corresponding to backwards movement are not present in this example.}
    \label{fig:stacked_images}
\end{figure}

The output of the equation is now a continuous value that is very sensitive to the movement of the phone, as it rotates around the vertical axis with each step. Figure~\ref{fig:stacked_images} shows the raw data computed from equation~\eqref{eq:motion}. The first two plots display the $ \omega_y $ component, whose values represent camera rotation between any two frames. Positive values indicate a rotation towards the left, and negative values towards the right, while values close to 0 indicated the camera has not rotated at all between two frames. The raw data has noisy values which can be smoothed out and processed into discrete values. We convert the values into the 8 classes that represent the rotation between frames. For visualization purposes, it can be considered that these classes correspond to a N/S/W/E compass, that also displays middle points between the cardinal directions. The bottom plot in Figure~\ref{fig:stacked_images} displays the sequence of classes along the frames of the video. A value of 0 indicates the class that represents the forward movement. The end user should in this situation see a compass displaying an arrow towards north.


\subsection{Model training}

\begin{figure}
    \centering
    \includegraphics[width=1\columnwidth]{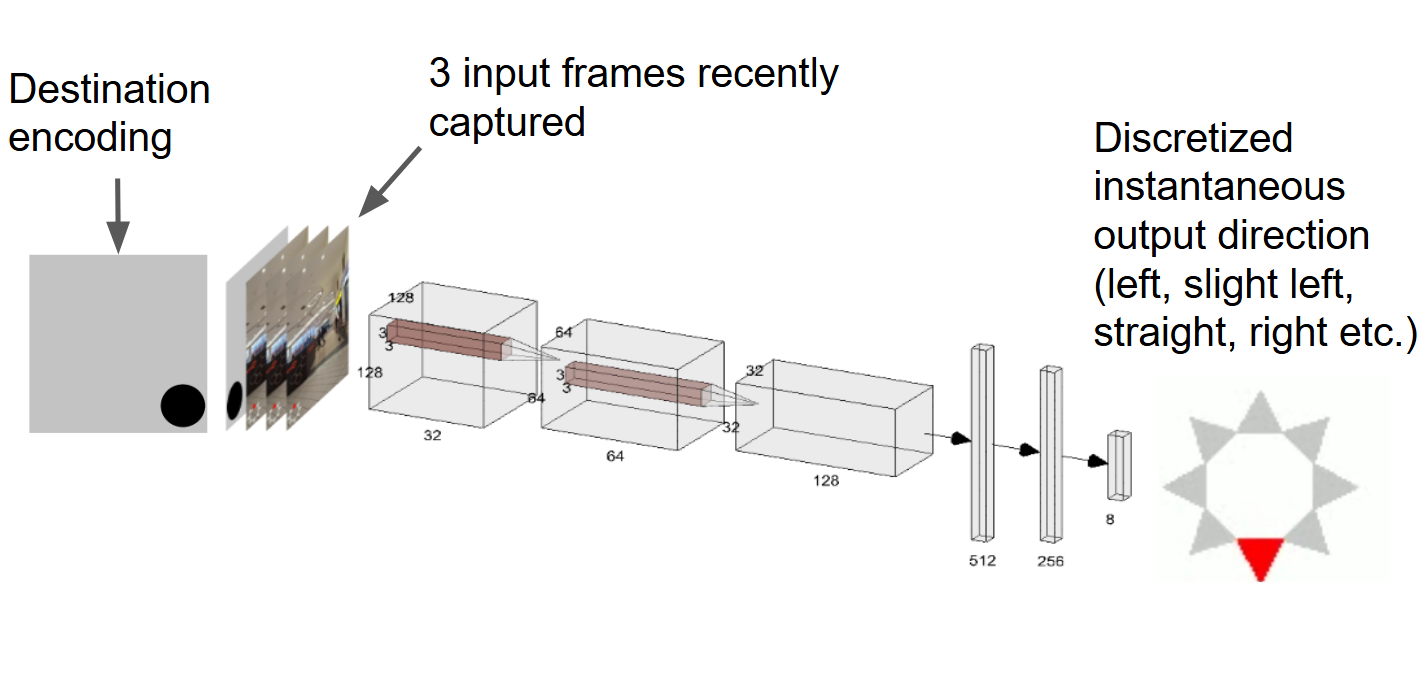}
    \caption{Proposed model architecture. \textbf{Input}: 3 frames from recorded footage of what the user sees while navigating and an additional encoded target image. \textbf{Model}: 8 million parameter CNN with 3 convolutional and 3 fully connected layers. \textbf{Output}: Next motion of the camera, discretized over 8 classes.  }
    \label{fig:arch}
\end{figure}

For our method, we experimented with deep ConvNets, as they are highly efficient, even in small size, for image 
classification~\cite{guo} and can be easily deployed on mobile devices. Our model architecture 
is shown in Fig.~\ref{fig:arch}. We found in tests that better results are obtained by using 3 RGB frames as input (current one, and two previous ones, 1 second apart). To encode the target destination, a black circle on a white background
whose center encodes the target is passed on one of the channels (note that the location encoding is decided in advance, but arbitrary, with no connection to the actual 2D location on the map).
Together with the 3 frames, we have a total of 10 channels at input. All images are resized to 128x128 for efficiency before being processed by the network. We also use standard data augmentation to randomly rotate the input images 15{\degree} in either direction, to account for the biased rotation a phone held in hand might have. The ground truth for the network is a one hot encoding of 8 classes which represent the predicted direction of walking from a given set of frames towards the target passed as an input.  
The model (Fig.~\ref{fig:arch}) consists of 3 convolutional layers and 3 fully connected layers, using only ReLU activations, totaling around 8 million learnable parameters. The output of the network is a probability for each of the 8 possible direction classes. For the results presented in this document, we use an Adam optimizer and train the model for at least 10 epochs with a learning rate of 0.001.
An important aspect of this task is the unbalanced nature that characterizes the data from large indoor scenes (as also observed in
Figure~\ref{fig:stacked_images}): while walking through a mall, or any large building with long hallways, most of the time a user will walk forwards, and only at times will they briefly take a turn to a different direction, if heading straight for a target. Due to this challenge, we chose to use a cross-entropy loss combined with sampling more training frames where the ground truth indicates a left or right turn. We also propose and test more advanced training procedures, to make the model
more robust to wrong turns, which we explain later in Sections \ref{sec:deeper_look}.

%% file: sec/4_experiments.tex
\section{Experiments} 

The goal of our experimental analysis is to show that the model can correctly associate image frames with walking directions,
at every time step (Sec. \ref{sec:every_time_step},
as well as at key moments in intersections or cases of turning around (Sections \ref{sec:key_moments}, \ref{sec:turn_around}). We took a more in depth-look at failure cases in Section \ref{sec:deeper_look} and proposed different masking strategies during training, which along with curriculum learning, addressed the failure cases in large part. Moreover, we also performed tests on out-of-distribution scenarios, and showed that the proposed method is robust to such distribution changes between testing and training cases.

Last but not least, we implemented our system on an Android Smart Phone and tested it in real time in the real world. The application always took the user to the correct place: the rare errors were usually immediately corrected by the system, while the user naturally disregarded directions given when distractors (moving people and obstacles) where covering large parts of the camera input. We will make the application, dataset and training code, publicly available. We also plan to have an extensive evaluation with many users and report their feedback and rate of success in the longer journal version, if accepted.

\subsection{Evaluation at every timestep}
\label{sec:every_time_step}

For a general understanding of the system's ability to learn to give directions from visual input, we first test its performance at every time step in unseen test paths (created from video segments unseen during training)
, by using three different evaluation measures: 1) Accuracy: it is the classical measure as percentage of correct decisions among all test cases (as compared to ground truth), along every test path, at every frame. 2) F1 Score: this is the standard F-measure, used such that the positive class includes all decisions that involve a change in direction and the negative class contain all test cases belonging to "move forward" action. The F1-measure, which is designed for problems with significant imbalance between positive and negative classes, is appropriate in this case, where changes in direction (the positive class) are rare compared to the "move forward" class. 3) The Angle Error measure, considers for each class the actual angle, as the center of the class bin around the circle of directions. Then the error is computed as the absolute difference in angle between the estimated direction and the ground truth. This measure is appropriate in order to give a sense of how far the model prediction is from ground truth: for example, a relatively small angle error (difference between right turn and slight-right turn) is less harmful than a large angle error of 180 degrees (right turn vs. left turn).

We test on different types of scenes, with different lengths and level of difficulty based on the number of people present (mostly empty vs. mostly crowded) and number of target destinations. We also tested different training strategies, as explained in the next section \ref{sec:deeper_look}, and present the results in \ref{tab:results2_corrected}. The results show that the presence of people can affect the performance, whereas the system is not sensitive to the length of the path or the number of possible destinations, which is very encouraging. Regarding the negative effect of the presence of people, it is handled by our automatic masking data augmentation strategies proposed and discussed in the next section (with performance presented in Table \ref{tab:results2_corrected}.

\begin{figure}[ht]
    \centering
    \includegraphics[width=1\columnwidth]{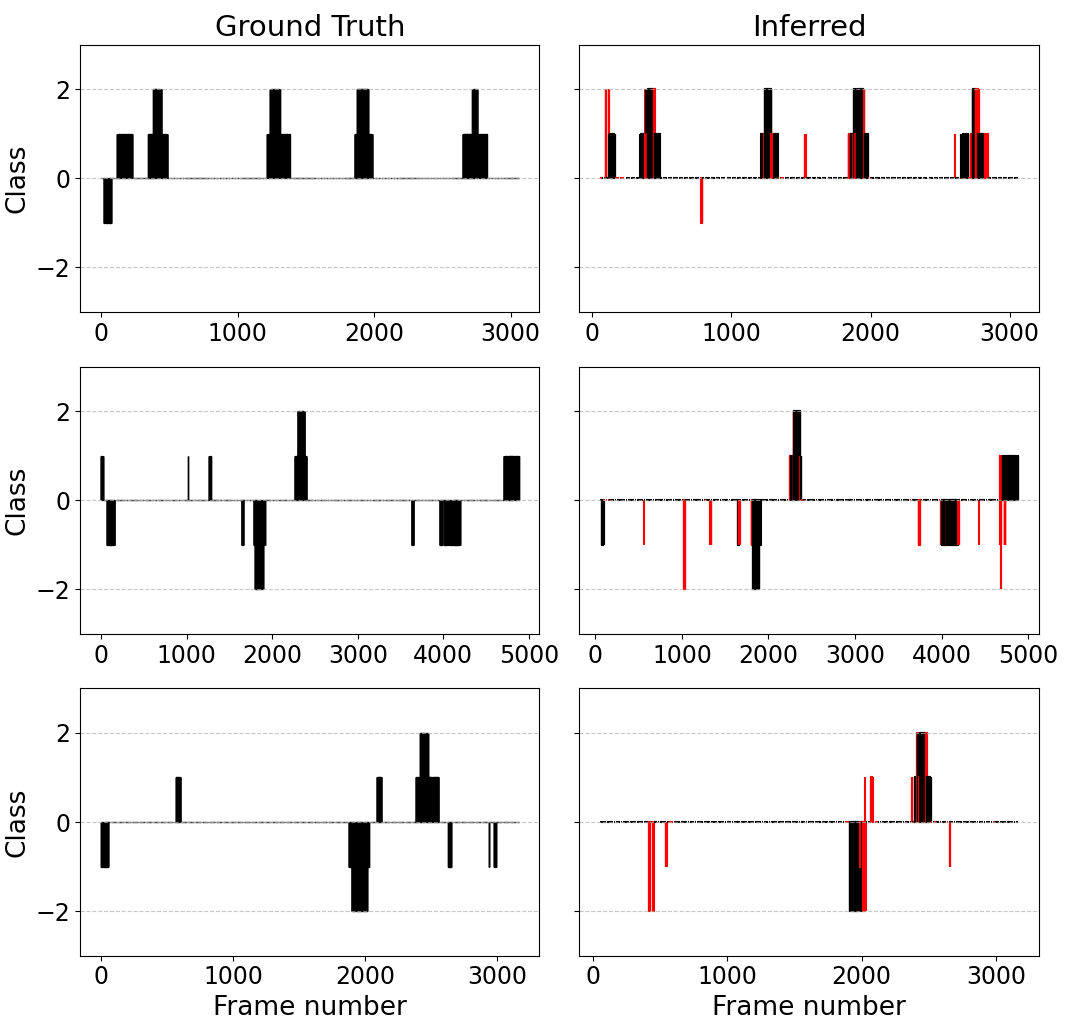}\\
    \caption{Comparison of real previously extracted motion data (\textbf{left}), compared to inferred values (\textbf{right}) for three different paths trained separately. Red areas denote wrongfully inferred classes for those frames.}
    \label{fig:stacked_images2}
\end{figure}

\begin{figure}[ht]
    \centering
    \includegraphics[width=1\columnwidth]{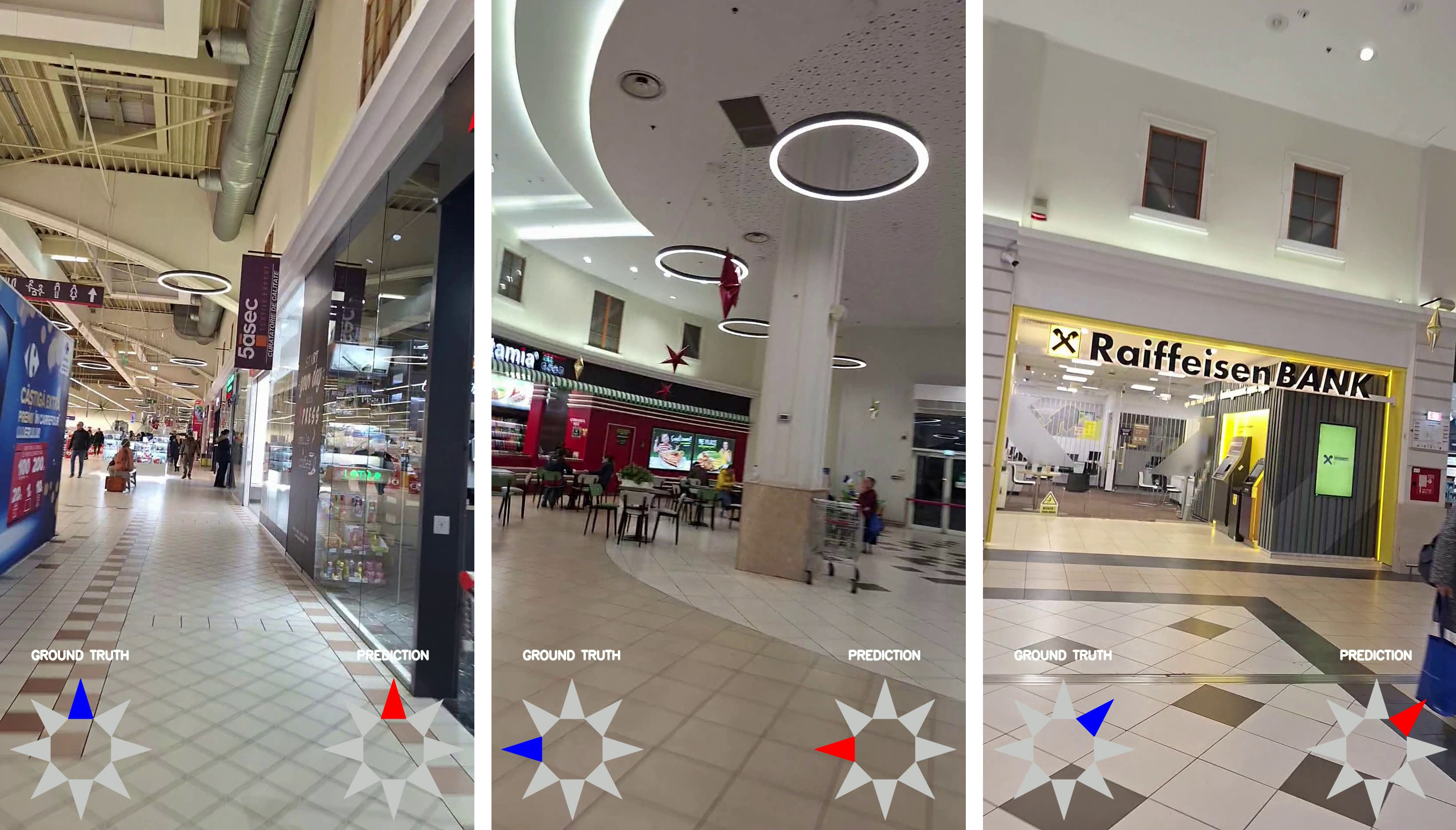}\\
    
    \caption{
        Examples of frames in our application, presenting the guidance compass twice, displaying the ground truth and the predicted value for those specific frames.
    }
    \label{fig:examples}
\end{figure}
\begin{table*}[ht]
\centering
\small
\caption{Comparison of accuracy, F1 scores, and angle error across various test scenarios, with row-wise averages (corrected).}
\label{tab:results2_corrected}
\begin{tabular}{lcc|cc|cc|cc}
\toprule
\textbf{Method}
  & \textbf{Crowded} & \textbf{Empty}
  & \textbf{Short paths}   & \textbf{Long paths}
  & \textbf{Single targets}  & \textbf{Many targets}
  & \textbf{Average} \\
\midrule
\multicolumn{8}{l}{\textit{Accuracy}} \\
\midrule
Baseline                  & 0.71 & 0.92 & 0.78 & 0.82 & 0.75 & 0.81 & 0.80 \\
PeopleMask                & 0.73 & 0.91 & 0.81 & 0.84 & 0.79 & 0.82 & 0.82 \\
RandMask                  & 0.73 & 0.88 & 0.79 & 0.85 & 0.83 & 0.84 & 0.82 \\
GradMask                  & 0.74 & 0.89 & 0.80 & \textbf{0.88} & 0.84 & 0.87 & 0.84 \\
GradMask + Curriculum     & \textbf{0.76} & \textbf{0.93} & \textbf{0.86} & 0.87 & \textbf{0.85} & \textbf{0.88} & \textbf{0.86} \\
\midrule
\multicolumn{8}{l}{\textit{Average F1 Score}} \\
\midrule
Baseline                  & 0.48 & 0.81 & 0.58 & 0.65 & 0.57 & 0.59 & 0.61 \\
PeopleMask                & 0.50 & 0.8 & 0.61 & 0.66 & 0.59 & 0.61 & 0.63 \\
RandMask                  & 0.47 & 0.76 & 0.62 & 0.67 & 0.59 & 0.60 & 0.62 \\
GradMask                  & 0.51 & 0.77 & 0.63 & \textbf{0.70} & 0.60 & 0.62 & 0.64 \\
GradMask + Curriculum     & \textbf{0.52} & \textbf{0.83} & \textbf{0.65} & 0.69 & \textbf{0.61} & \textbf{0.63} & \textbf{0.65} \\
\midrule
\multicolumn{8}{l}{\textit{Angle Error (degrees)}} \\
\midrule
Baseline                  & 14   &  7   & 11   & 10   & 12   & 12   & 11.00 \\
PeopleMask                & 13   &  8   &  9   &  9   & 10   & 11   & 10.00 \\
RandMask                  & 14   & 10   & 10   & 10   & 11   & 11   & 11.00 \\
GradMask                  & 13   &  8   &  \textbf{8}   & \textbf{8} & 10   &  8   & 9.17  \\
GradMask + Curriculum     & \textbf{12} & \textbf{6} &  \textbf{8} & 9    & \textbf{8} & \textbf{7} & \textbf{8.33}  \\
\bottomrule
\end{tabular}
\end{table*}

\subsection{Deeper look into model prediction}
\label{sec:deeper_look}

We notice that most failure cases take place when there are people right in front of the camera, which can easily explain why the predicted direction could be wrong. Humans also get confused, even in well-known scenes, when nearby obstacles (e,g. other moving people) are obstructing their view. During training, due to the presence of many people in videos, the system implicitly learns to base its output direction both on the final global navigation target and on the local navigation goal of avoiding collision with obstacles and moving people. 
However, when the moving people are too close to the camera, they occupy a large part of the input image and are expected to confuse the system. 

\begin{figure}
    \centering
    \includegraphics[width=0.8\columnwidth]{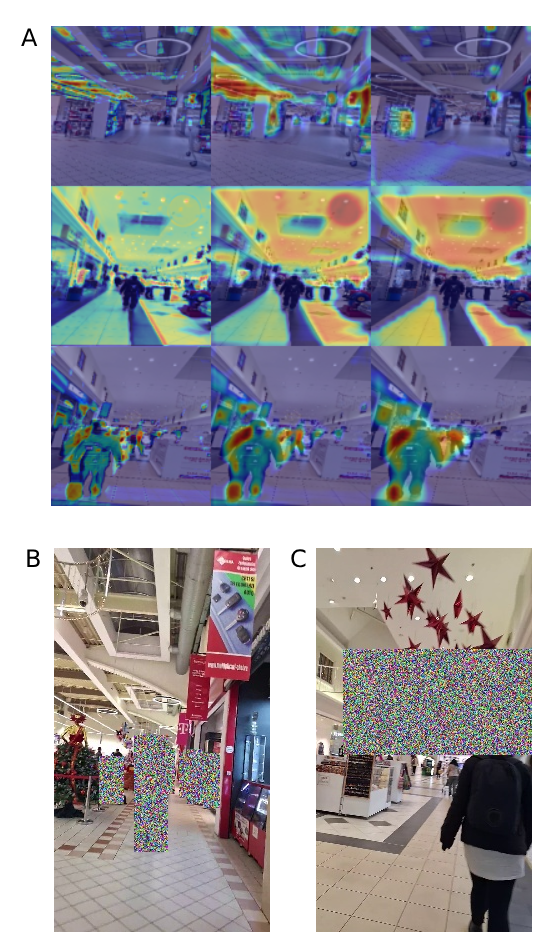}
    \caption{\textbf{A.} The network attention computed with Grad-CAM, among the first 3 convolutional layers. The top two rows show the attention when the correct direction prediction is moving forward - usually focused on the ceiling or the floor. The bottom row is the activation of a failure case, when the model predicts a right turn, due to a person being right in front of the camera.
    \textbf{B.} Augmentation based on people detection and removal. \textbf{C.} Cutout augmentation using distribution of network attention.}
    \label{fig:aug}
\end{figure}

For a more in-depth analysis, we use the guided backpropagation Grad-CAM algorithm\cite{ramprasaath, Springenberg}, that shows which parts of the input image are relevant for the final output of the model. Thus, we can
observe which image regions contribute the most to the system's wrong decisions.
In Fig.~\ref{fig:aug} we show the results of Grad-CAM (the input regions relevant for the final decision are
shown in stronger red), when making correct vs. wrong decisions. Visual inspection confirms the intuition: wrong decisions are mostly caused by people that are too close to the camera and occupy a relatively large part of the visual field.

These observations led us to design three different data augmentation procedures during training, in order to make the system more robust to such distracting situations (see Fig.~\ref{fig:aug}). First, we introduce the PeopleMask procedure in which we use a person detector\cite{Redmon} and fill the bounding boxes of people in the training set with random noise, and train the model again. With this modification, the network no longer associates direction with people, and instead focuses on static landmarks. Second, we test a random masking strategy (termed RandMask), in which masking boxes are chosen at random locations and sizes, similar to~\cite{Devries} - such a random strategy could make the system more robust by simply creating varied and much more difficult input, in which many details and structures, including people, are randomly cutout. 
In the third masking approach, we propose a method directly based on the output of the Grad-CAM attention, to make the input even harder during training. Termed GradMask, the idea is to randomly pick masking boxes (which again are filled with random noise) in the activation regions of Grad-CAM. This serves two relevant purposes: 1) it often covers the distracting people in front of the camera, obtaining the same effect as PeopleMask, without the need of using an additional off-the-shelf person detector; 2) it also covers relevant parts of the scene, similar to the second masking strategy, RandMask, thus leading to more robust training. We further extend the last GradMask strategy by adopting a curriculum learning strategy~\cite{soviany}, with a second-phase of training in which the model receives mostly hard training cases on which the model failed during the first learning phase.

The extensive experiments, on different type of scenes, in \ref{tab:results2_corrected}, present the performance of the different strategies and also confirm their effectiveness over the baseline (which is trained on data augmented with the graph-based path generation Algorithm 1).
Table~\ref{tab:results2_corrected} displays a detailed view of the results, comparing the chosen metrics across our three solutions (the baseline and two dataset augmentations). Generally, placing cutouts across the scene provides better results than covering the people directly, both outperforming the baseline training with no augmentations. 


\subsection{Evaluation at key moments in intersections}
\label{sec:key_moments}

In order to better understand the global capability of the system to guide the user to the correct target, we now look at key moments, in intersections and turn-around situations, where the model has to decide between several paths. If at every such key moment (intersection or turn-around) the right decision is made, then we expect the user to reach the correct target. Thus, we also test the system only on portions of the dataset where the camera approaches an intersection or when the user is facing the wrong way (results in Tab.~\ref{tab:results3}), for two scenarios: 1) when using original video footage along paths, without the training data augmentation using our graph-based path generation approach and 2) with the graph-based generated paths included in the training data: these are paths artificially created by concatenating segments from different videos, collected at different times, in order to create novel paths (not seen during actual collections), towards all possible targets.

\begin{table}[ht]
\centering
\caption{Average accuracy for intersections and turn–around, with and without our graph-based path generation (Algorithm 1). Note the highly significant performance boost produced by Alg. 1.}
\label{tab:results3}
\begin{adjustbox}{max width=\linewidth}
\begin{tabular}{lcc}
\toprule
\multirow{2}{*}{\textbf{Training setup}} & \multicolumn{2}{c}{\textbf{Accuracy}} \\
\cmidrule(lr){2-3}
 & \textbf{Intersection} & \textbf{Turning around} \\
\midrule
Without graph-based path generation  & 0.52 & 0.49 \\[4pt]
With graph-based path generation & \textbf{0.89} & \textbf{0.91} \\
\bottomrule
\end{tabular}
\end{adjustbox}
\end{table}

When using the graph-based path generation Algorithm 1, the training data is much richer, having virtually all possible paths included, from all starting points to all possible targets. These automatically generated paths, from real video segments, make the training set much larger and richer in terms of combinations of start and end points. Therefore, as the numbers show, the Algorithm is highly effective for improving global navigation performance.

\subsection{Evaluation of a special case: Turning around}
\label{sec:turn_around}

Sometimes, the target can be situated behind the user such that the best action is to turn around. Because our method naturally trains a model to predict the path of a camera, we tackle this case by training the model using the reversed video footage, such that the correct walking direction is in reverse, towards a specific target. Thus, if the user is ever facing the wrong way, the model will simply predict he must walk backwards, when they should turn around. We trained and tested such cases. As shown in Tab. ~\ref{tab:results3}, the model predicts with accuracy of 91\% the cases when the user should turn around. In practice we expect the effectiveness to be close to 100\% because when the system corrects itself after the user takes a few steps.

\subsection{Out-of-distribution testing}
\label{sec:out_of_distribution}

\begin{figure}
    \centering
    \includegraphics[width=1\columnwidth]{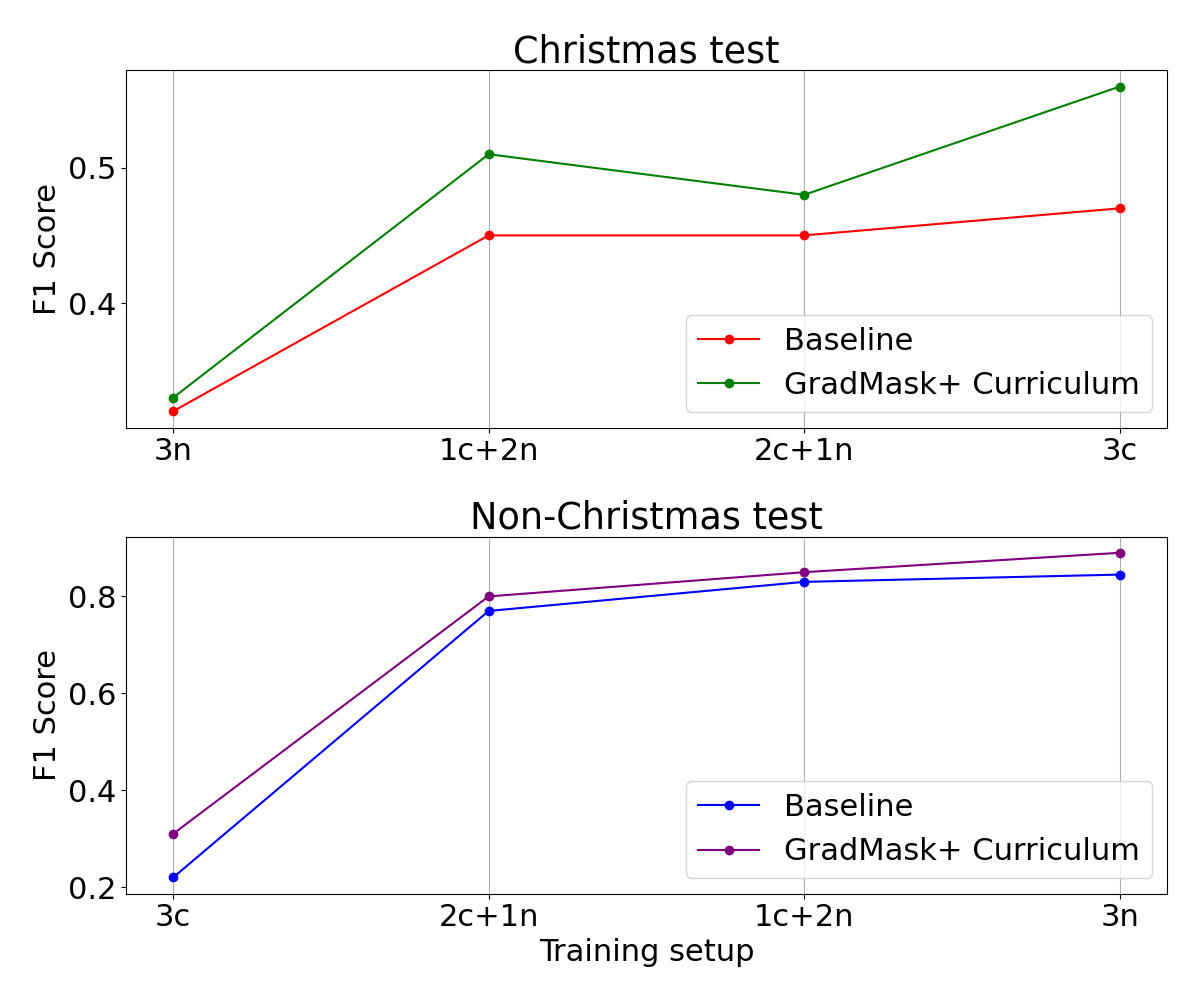}
    \caption{Out of distribution tests: two plots with F1 scores for testing during Christmas period (up) and before (non-) Christmas period (bottom) vs. different combinations of training videos, some filmed during Christmas (labeled as "c") and others filmed before Christmas (labeled as "n"). For example, 1c+2n on the x axis denotes a training set of 1 Christmas video and 2 non-Christmas ones. The visual distribution of scenes during Christmas is drastically different from the other ones, due to the abundance of seasonal decorations and number of people present. Each plot shows that as we increase the amount of in-distribution videos in the training set the performance increases drastically. Interestingly, a single in-distribution video is sufficient in both cases, to reach high performance, showing that the system is robust to a significant amount of out-of-distribution training data, as long as a minimum amount of in-distribution data is present during training. }
    \label{fig:distr}
\end{figure}

We further test our method in out-of-distribution cases. In practice the test data can often be captured during a different time of the year from the time of training data collection, so the visual appearance of scenes
could be significantly different between the training and testing periods, due to changes in interior decorations, shop facades, signs and people behavior. In Fig.~\ref{fig:distr} we present such experiments with videos collected during two distinct periods, Christmas and non-Christmas. We tested different combinations of train and test videos, with details explained in Fig.~\ref{fig:distr}.

%% file: sec/5_conclusions.tex
\section{Conclusions}

We presented a visual learning approach for indoor mapless navigation with a mobile device, in real time, to predict walking directions in complex, crowded indoor scenes. To our best knowledge, this is the first system of this kind, which, based on vision-only with a consumer-grade smart phone, demonstrates that accurate navigation can be achieved without
pre-installed infrastructure, special depth and other navigation sensors, or detailed scene maps. Our technical contributions include a novel graph-based path generation method for producing all possible paths from all start points to all targets, from limited amount of real video footage and different data augmentation techniques, including one based on an explainability model, Grad-CAM, which, combined with curriculum learning, makes the model robust to distractors (crowded scenes of many people walking in front of the camera). 

We implemented our system in an easy-to-use application for the Android system, which, along with our dataset, we plan to make publicly available, along with all our training code. Future work includes testing in different scenes and out-of-distribution scenarios as well as deploying our application to a larger number of users for a larger scale experimental evaluation.